\documentclass{article}
\usepackage{style}
\usepackage{times}
\usepackage{soul}
\usepackage{url}
\usepackage[hidelinks,draft]{hyperref}
\usepackage[utf8]{inputenc}
\usepackage[small]{caption}
\usepackage{subcaption}
\usepackage{graphicx}
\usepackage{amsmath}
\usepackage{amsthm}
\usepackage{booktabs}
\usepackage{soul}
\usepackage{amssymb}
\usepackage{listings}
\usepackage{setspace}
\usepackage[]{mdframed}
\usepackage{xspace}
\usepackage{color}
\usepackage[table]{xcolor}
\newtheorem{example}{Example}
\newtheorem{theorem}{Theorem}
\usepackage{xspace}
\usepackage{bbm} 
\usepackage{graphicx} 
\usepackage[shortlabels]{enumitem} 
\usepackage[font=footnotesize]{caption}
\usepackage{multirow,hhline} 
\usepackage{balance}
\usepackage{url}
\usepackage{pgf, tikz}
\usetikzlibrary{arrows, automata}
\usepackage{tabularx}
\usepackage{balance}
\newtheorem{definition}{Definition}
\newtheorem{proposition}{Proposition}

\newcommand{\wrt}{w.r.t.\xspace}
\renewcommand{\iff}{iff\xspace}

\def\A{\mbox{A}}

\def\L{\mathcal{L}}
\def\C{\mathcal{C}}
\def\W{\mathcal{W}}

\def\S{{\sigma}}

\newcommand{\Set}[1]{$S$}

\def\R{{\mbox{R}}}

\def\+{\mbox{+}}
\def\-{\mbox{-}}
\def\<{\langle}
\def\>{\rangle}

\newcommand{\sem}[1]{\ensuremath{\mathtt{#1}}}

\def\co{\tt co}
\def\st{\tt st}
\def\pr{\tt pr}
\def\ss{\tt sst}
\def\gr{\tt gr}

\renewcommand{\iff}{iff\xspace}

\def\styleex{\rm}

\def\L{{\mathcal{L}}}

\newcommand{\np}{\ensuremath{\textsc{N\!P}}}

\newcommand{\npc}{\np-c}

\newcommand{\conp}{{\normalfont co}\np}
\newcommand{\conpc}{\conp-c}

\newcommand{\pai}[1]{\ensuremath{\Pi_{#1}^{p}}}
\newcommand{\paic}[1]{\pai{#1}-c}

\newcommand{\sig}[1]{\ensuremath{\Sigma_{#1}^{p}}}
\newcommand{\sigc}[1]{\sig{#1}-c}

\newcommand{\ST}{\mathtt{st}}
\newcommand{\sst}{\mathtt{sst}}

\newcommand{\ldot}{\,{\bf .}\,}
\newcommand{\Def}{\textit{Def}}

\def\={\mbox{=}}
\def\Lab{\mbox{${\cal L}$}}

\def\i{\mbox{$\bf in$}}
\def\o{\mbox{$\bf  out$}}
\def\un{\mbox{$\bf und$}}

\usepackage{algorithm}
\usepackage[noend]{algorithmic}
\def\true{\mbox{$\tt true$}}
\def\false{\mbox{$\tt false$}}

\def\gr{\mbox{$\tt gr$}}
\def\co{\mbox{$\tt co$}}
\def\st{\mbox{$\tt st$}}
\def\pr{\mbox{$\tt pr$}}
\def\ss{\mbox{$\tt sst$}}

\usepackage{multirow}
\usepackage{rotating}
\usepackage{mathtools}
\usepackage{latexsym} 
\usepackage{hhline}
 \usepackage{braket}

\newcommand{\white}{\mbox{$\tt white$}}
\newcommand{\red}{\mbox{$\tt red$}}

\newcommand{\meat}{\mbox{$\tt meat$}}
\newcommand{\fish}{\mbox{$\tt fish$}}

\usepackage{amsmath}

\newcommand{\prob}[1]{\ensuremath{\mathsf{#1}}}
\newcommand{\noprob}[1]{\ensuremath{\text{#1}}}
\def\excp{\ensuremath{\prob{CF\text{-}EX}}}
\def\vecp{\ensuremath{\prob{CF\text{-}VE}}}
\def\cacp{\ensuremath{\prob{CF\text{-}CA}}}
\def\sacp{\ensuremath{\prob{CF\text{-}SA}}}
\def\exc{\ensuremath{\noprob{CF\text{-}EX}}}
\def\vec{\ensuremath{\noprob{CF\text{-}VE}}}
\def\cac{\ensuremath{\noprob{CF\text{-}CA}}}
\def\sac{\ensuremath{\noprob{CF\text{-}SA}}}

\def\exsp{\ensuremath{\prob{SF\text{-}EX}}}
\def\vesp{\ensuremath{\prob{SF\text{-}VE}}}
\def\casp{\ensuremath{\prob{SF\text{-}CA}}}
\def\sasp{\ensuremath{\prob{SF\text{-}SA}}}
\def\exs{\ensuremath{\noprob{SF\text{-}EX}}}
\def\ves{\ensuremath{\noprob{SF\text{-}VE}}}
\def\cas{\ensuremath{\noprob{SF\text{-}CA}}}
\def\sas{\ensuremath{\noprob{SF\text{-}SA}}}

\title{Counterfactual and Semifactual Explanations in Abstract Argumentation:\\  Formal Foundations,  Complexity and Computation}

\author{
Gianvincenzo Alfano\and
Sergio Greco\and
Francesco Parisi\and
Irina Trubitsyna \\
\affiliations
Department of Informatics, Modeling, Electronics and System Engineering,\\
University of Calabria, Italy\\
\emails
\{g.alfano, greco, fparisi, trubitsyna\}@dimes.unical.it
}

\begin{document}

\maketitle

\begin{abstract}
Explainable Artificial Intelligence and Formal Argumentation have received significant attention in recent years.  
Argumentation-based systems often 
lack explainability while supporting decision-making processes. 
Counterfactual and semifactual explanations are 
interpretability techniques that provide insights into the outcome of a model by generating alternative hypothetical instances. 
While there has been important work on counterfactual and semifactual explanations for Machine Learning models, 
less attention has been devoted to these kinds of problems in argumentation.
In this paper, we explore counterfactual and semifactual reasoning in abstract Argumentation Framework.
We investigate the computational complexity of counterfactual- and semifactual-based reasoning problems, showing that they 
are generally harder than classical argumentation problems such as credulous and skeptical acceptance.
Finally, we show that counterfactual and semifactual queries 
can be encoded in weak-constrained Argumentation Framework, 
and provide a computational strategy through ASP solvers. 
\end{abstract}

\section{Introduction}\label{sec:intro}

In the last decades, Formal Argumentation
has become an important research field in the area of 
knowledge representation and reasoning~\cite{Gabbay2021handbook}.
Argumentation has potential applications in several contexts, including e.g.
modeling dialogues, negotiation \cite{AmgoudDM07,DimopoulosMM19}, 
and persuasion~\cite{Prakken09}. 
Dung's Argumentation Framework (AF) is a simple yet powerful formalism for modeling disputes between two or more agents~\cite{Dung95}.
An AF consists of a set of \emph{arguments} and a binary \emph{attack} relation over the set of arguments that specifies the interactions between arguments:
intuitively, if argument $a$ attacks argument $b$, then $b$ is acceptable only if $a$ is not.
Hence, arguments are abstract entities whose status is entirely determined by the attack relation.
An AF can be seen as a directed graph, whose nodes represent arguments and edges represent attacks.
Several argumentation semantics---e.g. \textit{grounded} ($\gr$), \textit{complete} ($\co$), \textit{stable} ($\st$), \textit{preferred} ($\pr$), and \textit{semi-stable} ($\sst$)~\cite{Dung95,Caminada06}---have been defined for AF, leading to the characterization of $\S$-\textit{extensions}, that intuitively consist of the sets of arguments that can be collectively accepted under semantics $\S \in \{\gr,\co,\st,\pr,\sst\}$.

\begin{figure}[t!]
\centering
\includegraphics[scale=0.67]{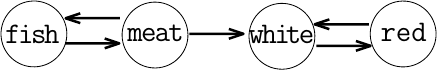}
\caption{AF $\Lambda$ of Example~\ref{running-example}.}\label{fig:intro}
\end{figure}

\begin{example}\label{running-example}\styleex
Consider the AF $\Lambda$ shown in Figure~\ref{fig:intro}, 
describing tasting menus proposed by a chef.
Intuitively, (s)he proposes to have either $\fish$ or $\meat$,
and to drink either $\white$ wine or $\red$ wine.  
However,  if serving $\meat$ then $\white$ wine is not paired with.  
AF $\Lambda$ has  three stable extensions 
(that are also preferred and semi-stable extensions): 
$E_1 = \{\fish,$ $\white\}$, 
$E_2 = \{\fish,$ $\red\}$,  and 
$E_3 =\{\meat,$ $\red \}$, 
representing alternative menus.~\hfill~$\Box$
\end{example}

Argumentation semantics can be also defined in terms of labelling~\cite{BaroniCG11}.
Intuitively, a $\sigma$-labelling for an AF is a total function $\L$ assigning to each argument the label $\i$ if its status is  accepted, $\o$ if its status is rejected, and  $\un$  if its status is undecided 
under semantics $\sigma$. 
For instance, the $\sigma$-labellings for AF $\Lambda$ of Example~\ref{running-example}, with $\sigma\in\{\st,\pr,\sst\}$, are as follows:
$\L_1 = \{\i(\fish), \o(\meat), \i(\white),\o(\red)\}$, \\
$\L_2 = \{\i(\fish), \o(\meat), \o(\white),\ \i(\red)\}$, \\
$\L_3 = \{\o(\fish), \ \i(\meat), \ \o(\white),\i(\red)\}$,\\
where $\L_i$ corresponds to extension $E_i$, with $i\in[1..3]$.

Integrating explanations in argumentation-based reasoners is important for enhancing argumentation and 
persuasion capabilities of software agents~\cite{MoulinIBD02,BexW16,CyrasBGTDTGH19,Miller19}.
For this reasons, several researchers explored how to deal with explanations in formal argumentation~(see related work in Section~\ref{sec:rw}).
Counterfactual and semifactual explanations are types of 
interpretability techniques that provide insights into the outcome of a model by generating hypothetical instances, 
known as counterfactuals and semifactual, 
respectively~\cite{kahneman1981simulation,mccloy2002semifactual}.
On one hand, a counterfactual explanation reveals what should have been different in an instance to obtain a diverse outcome~\cite{guidotti2022counterfactual}---minimum changes 
w.r.t. the given instance are usually considered~\cite{NIPS20}.
On the other hand, a semifactual explanation provides 
a maximally-changed instance yielding the same outcome of that considered~\cite{kenny2021generating}.

While there has been interesting work on counterfactual and semifactual explanations 
for ML models, e.g.~\cite{wu2019counterfactual,albini2020relation,romashov2022baycon,DandlCBB23,ijcai2023p732},
less attention has been devoted to these problems in  argumentation.

In this paper, we explore counterfactual and semifactual reasoning in AF.
In particular, in this context, 
counterfactual explanations help understanding 
what should have been different in a solution (a $\sigma$-labelling) provided by an argumentation-based system 
to have it approved by a user who is interested 
in the acceptance status of a goal argument.

\begin{example}\label{example:intro-counter}\styleex
Continuing with Example~\ref{running-example}, 
assume that the chef suggests the menu 
$\L_3=$ $\{\o(\fish),$ $\i(\meat),\o(\white), \i(\red)\}$ and the customer replies that (s)he likes everything except meat 
(as (s)he is vegetarian).
Therefore, the chef looks for the closest menu not containing 
$\meat$, that is
$\L_2=\{\i(\fish),$ $\o(\meat),$ $\o(\white), $ $\i(\red)\}$.
In this context, we say that $\L_2$ is a \textit{counterfactual} for $\L_3$ 
w.r.t. the goal argument $\meat$. ~\hfill{$\Box$}
\end{example}

Given a $\sigma$-labelling $\L$ of an AF $\Lambda$, 
and a goal argument $g$, 
a \textit{counterfactual} of $\L$ w.r.t. $g$ is 
a closest $\sigma$-labelling $\L'$ of $\Lambda$ 
that changes the acceptance status of $g$.
Hence, counterfactuals explain how to minimally change a solution 
to avoid a given acceptance status of a goal argument.
In contrast,   
semifactuals give the maximal changes to the considered solution in order to keep the status of a goal argument.
That is, 
a \textit{semifactual} of $\L$ w.r.t. goal $g$ is 
a farthest $\sigma$-labelling $\L'$ of $\Lambda$ 
that keeps the acceptance status of argument $g$.

\begin{example}\label{example:intro-semi}\styleex
Continuing with Example~\ref{running-example}, 
suppose now that a customer has tasted  menu $\L_3$ 
and asks to try completely new flavors while still 
maintaining the previous choice of wine as (s)he liked it a lot.
Here the chef is interested in the farthest menu containing 
$\red$ wine.
This menu is $\L_2=\{\i(\fish),\o(\meat),\o(\white), \i(\red)\}$. 
In this situation, we call $\L_2$ a \textit{semifactual} for $\L_3$ w.r.t. $\red$.~\hfill{$\Box$}
\end{example}

\noindent 
In this paper we introduce the concepts of counterfactual and semifactual explanations in AF. 
To the best of our knowledge, this is the first work addressing explainability queries in AF under both counterfactual and semifactual reasoning.

\vspace*{1mm}
\noindent 
\textbf{Contributions.}
Our main contributions are as follows.

\begin{itemize}
\item 
We introduce counterfactual-based and semifactual-based reasoning problems for existence, verification, and credulous and skeptical acceptance in AF.  
\item 
We investigate the complexity of above-mentioned problems
showing that they are generally harder than classical ones;
this particularly holds for verification 
and credulous acceptance problems.  
Notably, our results hold even for different generalizations of the concepts of counterfactual and semifactual,
of measures for computing the distance between labellings, and 
for multiple goals (cf. Section~\ref{sec:alternative}).
\item 
We show that the above-mentioned problems can be encoded through weak-constrained AF, that is a generalization of AF with both strong and weak constraints. 
\item
Finally, we provide an algorithm for the computation of counterfactuals and semifactuals which 
makes use of well-known ASP encoding of AF semantics. 
\end{itemize}

\section{Preliminaries}

In this section, after briefly recalling some complexity classes, 
we review the Dung's framework.

\subsection{Complexity Classes}
We recall the main complexity classes used in the  paper. 
PTIME (or simply $P$) consists of the problems that 
can be decided in polynomial-time.
Moreover, we have that\\ 
\noindent
$\bullet\Sigma_0^p = \Pi_0^p = \Delta_0^p = P$;\ \
$\bullet\Sigma_1^p = N\!P$ and $\Pi_1^p = coN\!P$;  and\\ 
\noindent
$\bullet\Delta_h^p\! =\! P^{\Sigma_{h-1}^p} $, 
$\Sigma_h^p\! =\! N\!P^{\Sigma_{h-1}^p}$,
and $ \Pi_h^p\! =\! co \Sigma_h^p$, $\forall h > 0$ \cite{book-Papadimitriou}. 
Thus, $P^C$ (resp., $N\!P^C$) denotes the class of problems that can be decided in polynomial time using an oracle in the class $C$ by a deterministic (resp., non-deterministic) Turing machine.
The class $\Theta_h^p = \Delta_h^p[log\ n]$ denotes the subclass of $\Delta_h^p$ consisting of the problems that can be decided in polynomial time by a deterministic Turing machine 
performing $O(log\ n)$ calls to an oracle in the class 
$\Sigma_{h-1}^p$.  It is known that $\Sigma_h^p\! \subseteq\!\Theta_{h+1}^p\! \subseteq \!\Delta_{h+1}^p \!\subseteq\! \Sigma_{h+1}^p \!\subseteq\! PSP\!AC\!E$ and   $\Pi_h^p \!\subseteq\!\Theta_{h+1}^p\! \subseteq\! \Delta_{h+1}^p\! \subseteq\! \Pi_{h+1}^p\! \subseteq\! PSP\!AC\!E$.

\subsection{Argumentation Framework}\label{sec:AF}

An abstract \textit{Argumentation Framework} (AF) is a pair $\<\A,\R\>$, 
where $\A$ is a set of \textit{arguments} and $\R \subseteq \A \times \A$ is a set of \textit{attacks}.
If $(a,b)\in \R$ then we say that $a$ attacks $b$.

Given an AF $\Lambda=\<\A,\R\>$  and a set $S \subseteq \A$ of arguments, 
an argument $a \in \A$ is said to be
\textit{i})	
\emph{defeated} w.r.t. $S$ iff $\exists b \in S$  such that 
$(b, a) \in  \R$, and 
\textit{ii})
\emph{acceptable} w.r.t. $S$ iff for every argument 
$b \in  \A$ with $(b, a) \in \R$, there is
$c \in S $ such that $(c,b) \in \R$.
The sets of defeated and acceptable arguments w.r.t. $S$ are
as follows (where $\Lambda$ is fixed):

\begin{itemize}
\item $\textit{Def}(S) = \{ a \in \A \mid \exists (b,a) \in \R \ldot b \in S
 \}$;
 \item
  $\textit{Acc}(S) = \{ a \in \A \mid \forall (b,a) \in \R  \ldot b \in \textit{Def}(S) \}$.
\end{itemize}

\noindent
Given an AF $\<\A,\R\>$, a set $S\subseteq \A$ of arguments is said to be 
\textit{i}) \emph{conflict-free} \iff $S \cap Def(S) = \emptyset$; 
\textit{ii})
 \emph{admissible} \iff it is conflict-free and $S \subseteq Acc(S)$.  

Different argumentation semantics have been proposed 
to characterize collectively acceptable sets of arguments, called \emph{extensions}~\cite{Dung95,Caminada06}.
Every extension is an admissible set  satisfying additional conditions.
Specifically, 
the complete, preferred, stable, semi-stable, and  grounded
extensions of an AF are defined as follows.

Given an AF $\<\A,\R\>$, 
a set $S \subseteq \A$ 
 is an \emph{extension} called:
 
\begin{itemize}\itemsep=-1pt
\item
\emph{complete} ($\tt co$) \iff it is an admissible set and $S = Acc(S)$;
\item
\emph{preferred} ($\tt pr$) \iff it is a $\subseteq$-maximal complete extension;
\item 
\emph{stable} ($\tt st$) \iff it is a total preferred extension, i.e. a preferred extension such that $S \cup Def(S) = \A$;
\item
\emph{semi-stable} ($\ss$) \iff it is a preferred extension such that $S \cup Def(S)$ is maximal $($\wrt $\subseteq)$;
\item
\emph{grounded} ($\tt gr$) \iff it is a $\subseteq$-minimal complete extension.
\end{itemize}

The argumentation semantics can be also defined in terms of \textit{labelling} \cite{BaroniCG11}.
A labelling for an AF $\< \A, \R \>$ is a total function $\Lab : \A \rightarrow \{ \i , \o, \un \}$ assigning to each argument a label:
$\Lab(a) = \i$  means that $a$ is accepted,
$\Lab(a) = \o$  means that $a$ is rejected, and 
$\Lab(a) = \un$ means that $a$ is undecided.

Let
$\i(\Lab)  = \{a \mid  a\in \A \wedge \Lab(a) = \i\}$,
$\o(\Lab) = \{a \mid  a\in \A \wedge \Lab(a) = \o\}$, and
$\un(\Lab)  = \{a \mid  a\in \A \wedge \Lab(a) = \un\}$,
a labelling $\Lab$ can be represented by means of a triple  $\< \i(\Lab), \o(\Lab),$ $ \un(\Lab)\>$.

Given an AF $\Lambda = \< \A, \R \>$,
a labelling $\Lab$ for $\A$ is said to be  
\emph{admissible (or legal)} if $\forall a \in \i(\Lab) \cup \o(\Lab)$ it holds that:\\
\noindent
(\textit{i})
$\Lab(a) = \o$ iff $\exists\, (b,a)\in \R$ such that $\Lab(b) = \i$; and\\
\noindent
(\textit{ii})
$\Lab(a)=\i$ iff  $\forall (b,a) \in \R$, $\Lab(b)=\o$ holds.

Moreover,
$\Lab$ is a \textit{complete} labelling (or $\co$-labelling) iff  conditions (\textit{i}) and (\textit{ii}) hold for all arguments $a \in \A$.

Between complete extensions and complete labellings there is a bijective mapping defined as follows:
for each extension $E$ there is a unique labelling
$\Lab(E) = \< E, \Def(E), \A \setminus (E \cup \Def(E))\>$
and for each labelling $\Lab$ there is a unique extension, that is $\i(\Lab)$.
We say that $\Lab(E)$ is the labelling \textit{corresponding} to $E$.
Moreover, we say that 
$\Lab(E)$ is a $\S$-labelling for a given AF $\Lambda$ 
and semantics
$\S \in $ $\{ \co, \pr, \ST, \sst,$ $ \gr\}$ 
iff $E$ is a $\S$-extension of $\Lambda$.  

In the following, we say that the \textit{status of an argument} $a$
w.r.t.\ a labelling $\Lab$ (or its corresponding extension $\i(\Lab)$)
is $\i$ (resp., $\o$, $\un$) iff
$\Lab(a) = \i$ (resp., $\Lab(a) = \o$, $\Lab(a) = \un$).
We will avoid to mention explicitly the labelling (or the extension) whenever it is understood.  

The set of complete (resp., preferred, stable, semi-stable, grounded) labellings of an AF $\Lambda$ will be denoted by $\co(\Lambda)$ (resp., $\pr(\Lambda)$, $\ST(\Lambda)$,  $\sst(\Lambda)$, $\gr(\Lambda)$).
All the above-mentioned semantics except the stable admit at least one labelling.
The grounded semantics, that admits exactly one labelling, is said to be a \emph{unique status} semantics, while the others are said to be \emph{multiple status} semantics.
With a little abuse of notation, in the following we also use $\gr(\Lambda)$
to denote the grounded labelling.
For any AF $\Lambda$,
it holds that: 
\textit{i}) 
$\ST(\Lambda) \subseteq \sst(\Lambda) \subseteq \pr(\Lambda) \subseteq \co(\Lambda)$, 
\textit{ii}) 
$\gr(\Lambda) \in \co(\Lambda)$, and
\textit{iii}) 
$\ST(\Lambda) \neq \emptyset$ implies that $\ST(\Lambda) = \sst(\Lambda)$.

For any pair ($\Lab$,$\Lab'$) of $\sigma$-labellings of AF $\Lambda=\<\A,\R\>$ (with $\sigma\in\{\tt gr,co,st,pr,sst\}$),   
we use $\delta(\Lab,\Lab')$ to denote the \textit{distance} 
$|\{a\in \A\mid \Lab(a)\neq \Lab'(a)\}|$ between 
$\Lab$ and $\Lab'$ in terms of the number of arguments 
having a different status.

\begin{figure}[t!]
\centering
\includegraphics[scale=0.5]{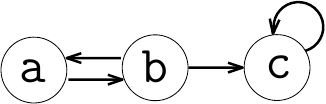}
\caption{AF of Example~\ref{ex:prel-af}.}\label{fig:prel-af}
\end{figure}

\begin{example}\label{ex:prel-af}\styleex 
Let $\Lambda=\< \A,\R\>$ be an AF where $\A=$ $\tt \{\tt a,$ $\tt b,$ $\tt c\}$ and $\R = \{\tt (a,b), $ $\tt (b,a),$ $\tt (b,c), $ $\tt (c,c) \}$ whose graph is shown in Figure~\ref{fig:prel-af}. 
AF $\Lambda$ has three complete labellings: $\L_1 = \{\un(\tt a),\un(\tt b),\un(\tt c)\}, \L_2 = \{\i(\tt a),\o(\tt b),\un(\tt c)\},$
and  $\L_3 = \{\o(\tt a),\i(\tt b),$ $\o(\tt c)\} $, 
and we have that  $\delta(\L_1,\L_2) = 2$ and  $\delta(\L_1,\L_3) = \delta(\L_2,\L_3) = 3$.
Moreover, the set of preferred labellings is $\pr(\Lambda)=\{ \L_2, \L_3 \}$, 
whereas the set of stable (and semi-stable) 
labellings is $\st(\Lambda)=\sst(\Lambda) = \{ \L_3 \}$, and the grounded labelling is $\L_1$.~\hfill$\Box$
\end{example}

Four canonical argumentation problems are \textit{existence}, \textit{verification}, and \textit{credulous} and 
\textit{skeptical acceptance}.
These problems can be formalized as follows.
Given an AF $\Lambda=\<\A,\R\>$,
for any semantics $\sigma\in\{\gr,\co,\st,\pr,\sst\}$,  
(\textit{i}) 
the existence problem (denoted $\mathsf{EX}^{{\S}}$) 
consists in deciding 
whether there is at least one $\sigma$-labelling for $\Lambda$; (\textit{ii}) 
the verification problem (denoted $\mathsf{VE}^{{\S}}$) consists in deciding whether a given labelling is a $\sigma$-labelling for $\Lambda$; 
and 
(\textit{iii})  given a (goal) argument $g\in \A$,
the \textit{credulous} (resp., \textit{skeptical}) 
acceptance problem, 
denoted as  $\mathsf{CA}^{\S}$ 
(resp., $\mathsf{SA}^{\S}$),
is the problem of deciding whether $\i(g)$ belongs to any (resp., all) 
$\S$-labellings of $\Lambda$.
Clearly, for the grounded semantics, which admits exactly one labelling, credulous and skeptical acceptance problems become identical.

The complexity of the above-mentioned problems has been
thoroughly investigated (see e.g.~\cite{Gabbay2021handbook} for a survey),  and the results are summarized in
Table~\ref{tab:my-table2}.

\section{Counterfactual Reasoning}

In this section, after formally defining the concept of counterfactual,
we investigate the complexity of counterfactual-based argumentation problems.

As stated next, 
a counterfactual of a given $\sigma$-labelling 
w.r.t. a given goal argument $g$ is 
a minimum-distance $\sigma$-labelling altering 
the acceptance status of $g$.

\begin{definition}[Counterfactual (CF)]\label{def:counter1}\rm
Let $\<\A,\R\>$ be an AF,  
$\sigma\in\{\tt gr,co,st,pr,sst\}$ a semantics,  
$g\in \A$ a goal argument, and 
$\Lab$ a $\sigma$-labelling for  $\<\A,\R\>$.
Then, a labelling $\Lab'\in \sigma(\<\A,\R\>)$ is a \textit{counterfactual} of $\Lab$ w.r.t.  $g$ 
if:
\begin{itemize}[leftmargin=20pt]
\item[$(i)$] 
$\Lab(g)\neq\Lab'(g)$, and 
 \item[$(ii)$]  there exists no $\Lab'' \in \sigma(\<\A,\R\>)$ 
 such that $\Lab(g)\neq\Lab''(g)$ and $\delta(\Lab,\Lab'')<\delta(\Lab,\Lab')$.
\end{itemize}
\end{definition}

We use ${\cal CF}^{\sigma}(g,\Lab)$
to denote the set of counterfactuals  of $\L$ w.r.t.   $g$.

\begin{example}\label{example:counter1}\styleex
Continuing with Example~\ref{example:intro-counter},
under stable semantics, 
for the labelling $\L_3 = \{\o(\fish), $ $\i(\meat), $ $\o(\white),$ $\i(\red)\}$, 
we have that $\L_2=\{\i(\fish),$ $\o(\meat),$ $\o(\white),$ $\i(\red)\}$ is its only counterfactual w.r.t. argument $\meat$,
as their distance, $\delta(\L_3,\L_2)=2$, is minimal. 
The other labelling 
$\L_1=\{\i(\fish),$ $\o(\meat),$ $\i(\white),$ $\o(\red)\}$, such that $\L_3(\meat)\neq\L_1(\meat)$ 
is not at minimum distance 
as $\delta(\L_3,\L_1)=4>\delta(\L_3,\L_2)$.
Therefore, ${\cal CF}^{\st}(\meat,\L_3)=\{\L_2\}$.~\hfill$\Box$
\end{example}

Observe that, in general,
the counterfactual relationship is not symmetric,
in the sense that $\L'\in {\cal CF}^{\S}(g,\L)$ does not 
entail that $\L\in {\cal CF}^{\S}(g,\L')$.
For instance, in our running example, we have that
$\L_3\in {\cal CF}^{\st}(\meat,\L_1)$ while $\L_1\not\in {\cal CF}^{\st}(\meat,\L_3)$.
Moreover,  counterfactual reasoning makes sense for multiple status semantics only.
In fact, for any AF $\<\A,\R\>$ and goal $g\in \A$,  
it holds that ${\cal CF}^{\gr}(g,\Lab)=\emptyset$. 
Thus, hereafter we focus on multiple semantics only,
as for the grounded semantics as all considered problems are trivial.

Finding a counterfactual means looking for a minimum distance labelling. 
The first problem we consider is a natural decision version 
of that problem.

\begin{definition}[CF-Existence Problem]\label{def:kexCF}\styleex
Let $\Lambda=\<\A,\R\>$ be an AF,   $\S\in\{\tt co,st,pr,sst\}$ a semantics,  $g\in \A$ a goal argument, 
$k\in \mathbb{N}$ an integer,  
and $\Lab\in \sigma(\Lambda)$ a labelling.  
Then,
$\excp^{\sigma}$ is the problem of deciding 
whether there exists a labelling $\Lab'\in \sigma(\Lambda)$ 
s.t.  $\L(g)\neq\L'(g)$ and $\delta(\Lab,\Lab') \leq k$.
\end{definition}

In the following, 
we use $\exc^{\sigma}_\Lambda(g,k,\Lab)$ 
(or simply $\exc^{\sigma}(g,k,\Lab)$ whenever $\Lambda$ is fixed)
to denote the output of the $\excp^{\sigma}$ problem 
with input $\Lambda, g, k$, and $\L$.

\begin{example}\label{example:counter2}\styleex
Continuing with Example~\ref{example:intro-counter},  assume the customer asks whether there is a menu not containing \meat\ and differing from menu $\L_3$ by at most two items.
Under stable semantics,  the answer to the question is given by 
$\exc^{\st}(\meat,2,\L_3)$: it is yes, as 
there is menu $\L_2\in\st(\Lambda)$,  with $\L_2(\meat)\neq\L_3(\meat)$ and $\delta(\L_3,\L_2)=2$.  
~\hfill{$\Box$}
\end{example}

The following theorem characterizes the complexity of the existence problem under counterfactual reasoning.

\begin{theorem}\label{thm:kexc}
$\excp^{\S}$ is:
\begin{itemize}
\item \np-complete for $\S\in\{\co,\st\}$; and
\item \sig{2}-complete for  $\S\in\{\pr,\sst\}$.
\end{itemize}
\end{theorem}

A problem related to $\excp^{\sigma}$ is that of 
verifying whether a given labelling $\L'$ is a counterfactual for $\L$ and $g$, and thus that the distance between the two labelling is minimal.

\begin{definition}[CF-Verification Problem]\label{def:ver-counter}\styleex
Let $\Lambda=\<\A,\R\>$ be an AF,   $\S\in\{\tt  co,st,pr,sst\}$ a semantics, $g\in \A$ a goal argument,
and $\Lab$  a $\sigma$-labelling of $\Lambda$.  
Then,  $\vecp^{\sigma}$ is the problem of deciding whether a given labelling $\Lab'$ belongs to ${\cal CF}^\sigma(g,\Lab)$.
\end{definition}

We use $\vec^{\sigma}_\Lambda(g,\L,\L')$ (or simply $\vec^{\sigma}(g,\L,\L')$) to denote the output of $\vecp^{\sigma}$ 
with input $\Lambda,g,\L,$ and $\L'$.
\begin{example}\label{example:counter2b}\styleex
Consider again the situation of Example~\ref{example:intro-counter},  and assume the customer is interested to know whether $\L_2$ is the closest menu to $\L_3$ not containing \meat.  
This problem can be answered by deciding $\vec^{\st}(\meat,\L_3,\L_2)$, which is true as we have that  ${\cal CF}^{\st}(\meat,\L_3)=\{\L_2\}$.~\hfill~$\Box$
\end{example}

Observe that 
$\vec^{\sigma}_\Lambda(g,\L,\L')$ is true 
iff 
\textit{i}) $\L(g)\neq\L'(g)$ 
\textit{ii}) 
$\L'\in \sigma(\Lambda)$ 
(i.e. the classical verification problem,
$\mathsf{VE}^{{\S}}$, is true for input $\L'$), and
\textit{iii})
$\exc^{\sigma}_\Lambda(g,\delta(\L,\L')-1,\L)$ is false.
Since the complexity of $\mathsf{VE}^{{\S}}$ is lower than that of 
$\exc^{\sigma}$ (cf. Table~\ref{tab:my-table2}), 
and checking $\L(g)\neq\L'(g)$ is in PTIME,
the above-mentioned three-steps strategy suggests that 
$\vec^{\sigma}$ can be decided by an algorithm in the 
complement complexity class of $\exc^{\sigma}$.
In fact, as entailed by the next result,
the problems $\vec^{\sigma}$ and $\exc^{\sigma}$ are on the same level of the polynomial hierarchy.

\begin{theorem}\label{theorem:counter-ver}
$\vecp^{\S}$ is:  
\begin{itemize}
\item \conp-complete for $\S\in\{\co,\st\}$; and 
\item \pai{2}-complete for  $\S\in\{\pr,\sst\}$.
\end{itemize}
\end{theorem}

Counterfactual-based acceptance problems extend credulous and skeptical reasoning w.r.t. a set ${\cal CF}^{\sigma}(g,\Lab)$
of counterfactuals for labelling $\L$ and goal $g$
(under $\sigma$).
They consist in deciding whether an additional argument $g'$ 
is accepted in any or all counterfactuals of $\L$ w.r.t. $g$, respectively.

\begin{definition}[CF-Acceptance Problems]\label{def:ver-counter}\styleex
Let $\Lambda=\<\A,\R\>$ be an AF,   $\S\in\{\tt  co,st,pr,sst\}$ a semantics,  $g\in \A$ an argument, 
and $\Lab$ a $\sigma$-labelling for $\Lambda$.
Then,  $\cacp^{\sigma}$ (resp., $\sacp^{\sigma}$) is the problem of checking whether a given goal argument $g'\in \A$ is accepted in any (resp., every) counterfactual in ${\cal CF}^{\sigma}(g,\Lab)$.
\end{definition}

We use $\cac^{\sigma}_\Lambda(g,\L,g')$ and $\sac^{\sigma}_\Lambda(g,\L,g')$ (or simply $\cac^{\sigma}(g,\L,g')$ and $\sac^{\sigma}(g,\L,g')$) to denote
the output of the $\cacp^{\sigma}$ and $\sacp^{\sigma}$ 
with input $\Lambda, g,\L$ and $g'$, respectively.

\begin{example}\label{example:counter3}\styleex
Consider Example~\ref{example:intro-counter}, and assume the customer is interested to know whether in any (resp., all) of the closest menus to $\L_3$ not containing \meat, 
\red\ wine is present. 
Both answers are positive as $\cac^{\st}(\meat,$ $\L_3,\red)$ and $\sac^{\st}(\meat,\L_3,\red)$ are both true,
since ${\cal CF}^{\st}(\meat,\L_3)=\{\L_2\}$ and 
$\L_2(\red)=\i$.~\hfill{$\Box$}
\end{example}

Our next results address the complexity of 
counterfactual-based credulous and skeptical acceptance.

\begin{theorem}\label{theorem:counter-ca}
$\cacp^{\S}$ is: 
\begin{itemize}
\item \np-hard\ and in $\Theta_2^p$ for $\sigma\in\{\co,\st\}$; and 
\item \sig{2}-hard\ and in $\Theta_3^p$ for $\sigma\in \{\pr,\sst\}$.
\end{itemize}
\end{theorem}

\begin{theorem}\label{theorem:counter-sa}
$\sacp^{\S}$ is: 
\begin{itemize}
\item \conp-hard\ and in $\Theta_2^p$ for $\sigma\in\{\co,\st\}$; and 
\item \pai{2}-hard\ and in $\Theta_3^p$ for $\sigma\in \{\pr,\sst\}$.
\end{itemize}
\end{theorem}

\begin{table*}[t!]
\footnotesize
\begin{center}
\begin{tabular}{ccccccccccccccc}
  \cline{2-13} 
  \multicolumn{1}{c|}{} &
  \multicolumn{4}{c||}{Classical problems} &
  \multicolumn{4}{c||}{Counterfactual-based problems} &
  \multicolumn{4}{c||}{Semifactual-based problems} 
  \\ \hline
\multicolumn{1}{|c||}{$\S$} &
  \multicolumn{1}{c|}{\!\!\!$\mathsf{EX}^{{\S}}$\!\!\!} &
  \multicolumn{1}{c|}{$\mathsf{VE}^{{\S}}$} &
  \multicolumn{1}{c|}{$\mathsf{CA}^{{\S}}$} &
    \multicolumn{1}{c||}{$\mathsf{SA}^{{\S}}$} &
\multicolumn{1}{c|}{\!\!\!$\excp^{{\S}}$\!\!\!} &
  \multicolumn{1}{c|}{\!\!\!$\vecp^{{\S}}$\!\!\!} &
  \multicolumn{1}{c|}{$\cacp^{{\S}}$} &
  \multicolumn{1}{c||}{$\sacp^{{\S}}$}   &
  \multicolumn{1}{c|}{\!\!\!$\exsp^{{\S}}$\!\!\!} &
  \multicolumn{1}{c|}{\!\!\!$\vesp^{{\S}}$\!\!\!} &
  \multicolumn{1}{c|}{$\casp^{{\S}}$} &
  \multicolumn{1}{c||}{$\sasp^{{\S}}$} 
\\ \hline \hline
   \multicolumn{1}{|c||}{$\sem{co}$} &
  \multicolumn{1}{c|}{T} & 
  \multicolumn{1}{c|}{P} &
 \multicolumn{1}{c|}{\npc} &
  \multicolumn{1}{c||}{{P}} &
     \multicolumn{1}{c|}{\cellcolor{cyan!20}{\npc}} & 
  \multicolumn{1}{c|}{\cellcolor{cyan!20}{\conpc}} &
   \multicolumn{1}{c|}{\cellcolor{cyan!20}{\np-h, $\Theta_2^p$}} &
   \multicolumn{1}{c||}{\cellcolor{cyan!20}{\conp-h, $\Theta_2^p$}} &
     \multicolumn{1}{c|}{\cellcolor{cyan!20}\npc} & 
  \multicolumn{1}{c|}{\cellcolor{cyan!20}{\conpc}} &
   \multicolumn{1}{c|}{\cellcolor{cyan!20}{{\np-h,} $\Theta_2^p$}} &
   \multicolumn{1}{c||}{\cellcolor{cyan!20}{{\conp-h,} $\Theta_2^p$}}
   \\ \hline
   \multicolumn{1}{|c||}{$\sem{st}$} &
  \multicolumn{1}{c|}{\!\!\!\npc\!\!\!} & 
  \multicolumn{1}{c|}{{P}} &
 \multicolumn{1}{c|}{\npc} &
  \multicolumn{1}{c||}{\conpc} &
     \multicolumn{1}{c|}{\cellcolor{cyan!20}{\npc}} & 
  \multicolumn{1}{c|}{\cellcolor{cyan!20}{\conpc}} &
   \multicolumn{1}{c|}{\cellcolor{cyan!20}{\np-h, $\Theta_2^p$}} &
   \multicolumn{1}{c||}{\cellcolor{cyan!20}{\conp-h, $\Theta_2^p$}} &
     \multicolumn{1}{c|}{\cellcolor{cyan!20}\npc} & 
  \multicolumn{1}{c|}{\cellcolor{cyan!20}{\conpc}} &
   \multicolumn{1}{c|}{\cellcolor{cyan!20}{{\np-h,} $\Theta_2^p$}} &
   \multicolumn{1}{c||}{\cellcolor{cyan!20}{{\conp-h,} $\Theta_2^p$}}
   \\ \hline
   \multicolumn{1}{|c||}{$\sem{pr}$} &
  \multicolumn{1}{c|}{T} & 
  \multicolumn{1}{c|}{\!\!\!\conpc\!\!\!} &
 \multicolumn{1}{c|}{\npc} &
  \multicolumn{1}{c||}{\paic{2}} &
     \multicolumn{1}{c|}{\cellcolor{cyan!20}{\sigc{2}}} & 
  \multicolumn{1}{c|}{\cellcolor{cyan!20}{\paic{2}}} &
   \multicolumn{1}{c|}{\cellcolor{cyan!20}{\sig{2}-h, $\Theta_3^p$}} &
   \multicolumn{1}{c||}{\cellcolor{cyan!20}{\pai{2}-h, $\Theta_3^p$}} &
     \multicolumn{1}{c|}{\cellcolor{cyan!20}{\sigc{2}}} & 
  \multicolumn{1}{c|}{\cellcolor{cyan!20}{\paic{2}}} &
   \multicolumn{1}{c|}{\cellcolor{cyan!20}{{\sig{2}-h, }$\Theta_3^p$}} &
   \multicolumn{1}{c||}{\cellcolor{cyan!20}{{\pai{2}-h, }$\Theta_3^p$}}
   \\ \hline
   \multicolumn{1}{|c||}{$\sem{sst}$} &
  \multicolumn{1}{c|}{T} & 
  \multicolumn{1}{c|}{\!\!\!\conpc\!\!\!} &
 \multicolumn{1}{c|}{\sigc{2}} &
  \multicolumn{1}{c||}{\paic{2}} &
     \multicolumn{1}{c|}{\cellcolor{cyan!20}{\sigc{2}}} & 
  \multicolumn{1}{c|}{\cellcolor{cyan!20}{\paic{2}}} &
   \multicolumn{1}{c|}{\cellcolor{cyan!20}{\sig{2}-h, $\Theta_3^p$}} &
   \multicolumn{1}{c||}{\cellcolor{cyan!20}{\pai{2}-h, $\Theta_3^p$}} &
     \multicolumn{1}{c|}{\cellcolor{cyan!20}{\sigc{2}}} & 
  \multicolumn{1}{c|}{\cellcolor{cyan!20}{\paic{2}}} &
   \multicolumn{1}{c|}{\cellcolor{cyan!20}{{\sig{2}-h, }$\Theta_3^p$}} &
   \multicolumn{1}{c||}{\cellcolor{cyan!20}{{\pai{2}-h, }$\Theta_3^p$}}
   \\ \hline
\end{tabular}
\caption{Complexity of classical, counterfactual-based and semifactual-based problems in AF under complete ($\tt co$), stable ($\tt st$), preferred ($\tt pr$), and 
semi-stable ($\tt sst$) semantics.
For any complexity class $C$, 
$C$-c (resp., $C$-h) means $C$-complete (resp., $C$-hard); an interval $C$-h, $C'$ means $C$-hard and in $C'$.  
New results are highlighted in cyan. 
T means trivial (from the computational standpoint).\label{tab:my-table2}
}
\end{center}
\end{table*}

\section{Semifactual Reasoning}\label{sec:expressiv}

In this section, following what is done in the previous section, we investigate the complexity of semifactual-based argumentation problems.
We first introduce the concept of semifactual 
that, in a sense, 
is symmetrical and complementary to that of a counterfactual.

\begin{definition}[Semifactual (SF)]\label{def:semi1}\rm
Let $\<\A,\R\>$ be an AF,  
$\sigma\in\{\tt gr,co,st,pr,sst\}$ a semantics,  
$g\in \A$ a goal argument, and 
$\Lab$ a $\sigma$-labelling for  $\<\A,\R\>$.
Then, $\Lab'\in \sigma(\<\A,\R\>)$ is a semifactual of $\Lab$ w.r.t.  $g$ 
if:
\begin{itemize}[leftmargin=20pt]
\item[$(i)$] 
$\Lab(g)=\Lab'(g)$, and 
 \item[$(ii)$]  there exists no $\Lab'' \in \sigma(\<\A,\R\>)$ 
 such that  $\Lab(g)=\Lab''(g)$ and $\delta(\Lab,\Lab'')>\delta(\Lab,\Lab')$.
\end{itemize}
\end{definition}

We use  ${\cal SF}^{\sigma}(g,\Lab)$
to denote the set of semifactuals of $\L$ w.r.t. $g$.

\begin{example}\label{example:semi1}\styleex
Consider the stable labelling
$\L_3 = \{\o(\fish), \i(\meat), \o(\white),\i(\red)\}$ 
for the AF of Example~\ref{example:intro-semi}.  
We have that $\L_2=\{\i(\fish),$ $\o(\meat),$ $\o(\white),$ $\i(\red)\}$ is the only semifactual of $\L_3$ w.r.t. $\red$
as there is no other labelling agreeing on $\red$ and having 
distance greater than $\delta(\L_3,\L_2)=2$. 
In fact, while $\L_1=\{\i(\fish),$ $\o(\meat),$ $\i(\white),$ $\o(\red)\}$ have distance $\delta(\L_3,\L_1)=4$, 
it is not a semifactual for $\L_3$ w.r.t.  $\red$ 
as $\L_1(\red)\neq\L_3(\red)$.
Thus, ${\cal SF}^{\st}(\red,\L_3)=\{\L_2\}$.~\hfill{$\Box$}
\end{example}

Similarly to the case of counterfactuals, 
the semifactual relationship is not symmetric, that is, 
$\L'\in {\cal SF}^{\S}(g,\L)$ does not 
entail that $\L\in {\cal SF}^{\S}(g,\L')$.

As for the case of counterfactuals,
 semifactual reasoning makes sense only for multiple status semantics.
Thus,  hereafter,  we do not consider the grounded semantics.

The semifactual-based existence problem is as follows.

\begin{definition}[SF-Existence Problem]\label{def:kex}\styleex
Let $\Lambda=\<\A,\R\>$ be an AF,   $\S\in\{\tt  co,st,pr,sst\}$ a semantics,  $g\in \A$ a goal argument, 
$k\in \mathbb{N}$ an integer,  
and $\Lab\in \sigma(\Lambda)$ a labelling.  
Then,
$\exsp^{\sigma}$ is the problem of checking whether there exists a  labelling $\Lab'\in \sigma(\Lambda)$ s.t.  $\L(g)=\L'(g)$ and $\delta(\Lab,\Lab') \geq k$.
\end{definition}

We use $\exs^{\sigma}_\Lambda(g,k,\Lab)$ (or simply $\exs^{\sigma}(g,k,\Lab)$ whenever $\Lambda$ is fixed) to denote 
the output of $\exsp^{\sigma}$ with input 
$\Lambda, g,k,$ and $\L$.

\begin{example}\label{example:semi2}\styleex
Continuing with Example~\ref{example:intro-semi},  assume the customer is interested to know whether there exists a menu containing \red\ wine and differing from $\L_3$ by at least two items.
Under stable semantics,  the answer to this question is yes,
as there exists  
menu $\L_2\in\st(\Lambda)$, with $\L_2(\red)=\L_3(\red)$  and $\delta(\L_3,\L_2)=2$,  i.e.  $\exs^{\st}(\red,2,\L_3)$ is true.~\hfill{$\Box$}
\end{example}

The following theorem characterizes the complexity 
of the existence problem under semifactual reasoning.

\begin{theorem}\label{thm:kexs}
$\exsp^{\S}$ is:
\begin{itemize}
\item \np-complete for $\S\in\{\co,\st\}$; and 
\item \sig{2}-complete for  $\S\in\{\pr,\sst\}$.
\end{itemize}
\end{theorem}

For semifactuals, the verification problem is checking 
whether a given labelling $\L'$ is a semifactual for $\L$ and $g$
(hence the distance between the two labelling is maximal).

\begin{definition}[SF-Verification Problem]\label{def:ver-semi}\styleex
Let $\Lambda=\<\A,\R\>$ be an AF,   $\S\in\{\tt  co,st,pr,sst\}$ a semantics,  $g\in \A$ a goal argument,
and $\Lab$  a $\sigma$-labelling of $\Lambda$.  
Then,  $\vesp^{\sigma}$ is the problem of checking whether a given labelling $\Lab'$ belongs to ${\cal SF}^\sigma(g,\Lab)$.
\end{definition}

We use $\ves^{\sigma}_\Lambda(g,\L,\L')$ (or simply $\ves^{\sigma}(g,\L,\L')$) to denote the output of $\vesp^{\sigma}$  with 
input $\Lambda,g,\L,$ and $\L'$.

\begin{example}\label{example:semi2b}\styleex
Consider the situation in Example~\ref{example:intro-semi}, 
and assume the customer is interested to know whether $\L_2$ is the farthest menu w.r.t. $\L_3$ containing \red\ wine.  
This problem can be answered by deciding $\ves^{\st}(\meat,\L_3,\L_2)$, which is true as we have that  ${\cal SF}^{\st}(\red,\L_3)=\{\L_2\}$.~\hfill~$\Box$
\end{example}

\begin{theorem}\label{theorem:semi-ver}
$\vesp^{\S}$ is:  
\begin{itemize}
\item
\conp-complete for $\S\in\{\co,\st\}$; and 
\item 
\pai{2}-complete\ for  $\S\in\{\pr,\sst\}$.
\end{itemize}
\end{theorem}

Finally, we investigate semifactual-based acceptance problems,
that is credulous and skeptical reasoning 
w.r.t. a set ${\cal SF}^{\sigma}(g,\Lab)$ of semifactual for labelling $\L$ and goal $g$.

\begin{definition}[SF-Acceptance Problems]\label{def:ver-semi}\styleex
Let $\Lambda=\<\A,\R\>$ be an AF,   $\S\in\{\tt  co,st,pr,sst\}$ a semantics,  $g\in \A$ a goal argument,  
and $\Lab$ a $\sigma$-labelling for $\Lambda$.
Then,  $\casp^{\sigma}$ (resp., $\sasp^{\sigma}$) is the problem of deciding whether a given goal argument $g'\in \A$ is accepted in any (resp., every) semifactual 
$\Lab'\in {\cal SF}^{\sigma}(g,\Lab)$.
\end{definition}

We use $\cas^{\sigma}_\Lambda(g,\L,g')$ and $\sas^{\sigma}_\Lambda(g,\L,g')$ (or $\cas^{\sigma}(g,\L,g')$ and $\sas^{\sigma}(g,\L,g')$) to denote 
the output of $\casp^{\sigma}$ and $\sasp^{\sigma}$ with 
input $\Lambda, g,\L$ and $g'$, respectively.

\begin{example}\label{example:semi3}\styleex
In our running example, assume the customer is interested to know whether in any (resp., all) of the farthest menus w.r.t.  $\L_3$ containing \red\ wine,  \fish\ is present. 
Both answers are positive as  
$\cas^{\st}(\red,\L_3,\fish)$ and $\sas^{\st}(\red,\L_3,\fish)$ are true. ~\hfill{$\Box$}
\end{example}

The next theorems state the complexity of 
semifactual-based credulous and skeptical  acceptance problems.

\begin{theorem}\label{theorem:semi-ca}
$\casp^{\S}$ is: 
\begin{itemize}
\item {\np-hard\ and} in $\Theta_2^p$ for $\sigma\in\{\co,\st\}$; and 
\item {\sig{2}-hard\ and} in $\Theta_3^p$ for $\sigma\in \{\pr,\sst\}$.
\end{itemize}
\end{theorem}

\begin{theorem}\label{theorem:semi-sa}
$\sasp^{\S}$ is: 
\begin{itemize}
\item {\conp-hard\ and} in $\Theta_2^p$ for $\sigma\in\{\co,\st\}$; and 
\item {\pai{2}-hard}\ and in $\Theta_3^p$ for $\sigma\in \{\pr,\sst\}$.
\end{itemize}
\end{theorem}

Thus, semifactual- and counterfactual-based reasoning problems share the same complexity bounds.
Moreover, the considered problems are generally harder than the corresponding classical argumentation problems; this particularly holds if we focus on the verification 
problem and credulous acceptance under preferred semantics.

\section{WAF and ASP Mappings}
We first show that counterfactual and semifactual explanations  
can be encoded through Weak constrained AF (WAF), which is a generalization of AF with strong and weak constraints, and then provide an algorithm for computing counterfactuals and 
semifactuals by making use of well-known ASP encoding of AF semantics as well as of constraints
capturing counterfactuals and semifactuals' semantics.

 \subsection*{Weak Constrained AF}\label{sec:WAF}
 Constraints in argumentation frameworks have been investigated in several works \cite{Coste-MarquisDM06,Arieli15,SakamaS20,AlfanoGPT21a}.
They extend AF by considering a set of strong and weak constraints, 
that are sets of propositional formulae to be satisfied by extensions.
Intuitively, constraints introduce subjective knowledge of agents, 
whereas the AF encodes objective knowledge.    
Strong constraints in AF allow restricting the set of feasible solutions, but do not help in finding ``best'' or preferable solutions. 
To express this kind of conditions,   \textit{weak} constraints have been introduced, that is, constraints that are required to be satisfied 
\textit{if possible}~\cite{AlfanoGPT21a}. 
In the following, for the sake of presentation,
we consider constraints as propositional formulae built over labelled arguments, as e.g. in~\cite{SakamaI00}, instead of propositional formulae defined over argument literals.
An AF with strong and weak constraints is said to be a \textit{Weak constrained Argumentation Framework} (WAF).

\begin{definition}[WAF]\label{def:waf}
A \emph{Weak Constrained AF (WAF)} is a quadruple {\emph{$\<\A, \R, \C,\W\>$}} where {\emph{$\<\A, \R\>$}} is an AF,  
and 
$\C$ and $\W$ are sets of 
propositional formulae called \emph{strong} and \emph{weak}  constraints, respectively,
both built from the set of labeled arguments 
$\lambda_A=\{\i(a),\o(a),\un(a)\mid a\in {\emph{\A}}\}$ by using the connectives $\neg$, $\vee$, and $\wedge$.
\end{definition}

Observe that atoms in $\lambda_A$ represent the possible labels 
for each argument in $A$.
Given a labelling $\L$ and an argument $a$,  
we use $\ell(\L,a)$ to denote the atom in $\lambda_A$ reflecting the label of $a$ 
w.r.t. $\L$, that is,
$\i(a)$ (resp., $\un(a)$, $\o(a)$) if $\L(a)=\i$ (resp., $\L(a)=\un$, $\L(a)=\o$). 

We say that a labelling $\L$ satisfies a constraint $\kappa$ if and
only if $M(\L)=\{\ell(\L,a)\, |\, a \in A\}$ is a (2-valued) model of $\kappa$, denoted as $\L \models \kappa$, that is,  
if the formula obtained from $\kappa$ by replacing every atom occurring in $M(\L)$ with $\true$, and every atom not occurring in $M(\L)$ with $\false$, evaluates to true.
Moreover, we say that $\L$ satisfies a set $\{\kappa_1,\dots,\kappa_n\}$ of constraints, denoted as $\L\models \{\kappa_1,\dots,\kappa_n\}$, whenever $\L \models \kappa_i\  \forall i\in [1,n]$.
Maximum-cardinality semantics for WAF prescribes as preferable extensions 
those satisfying the largest number of weak constraints~\cite{AlfanoGPT21a}.
This is similar to the semantics of weak constraints in  
DLV~\cite{AlvianoCDFLPRVZ17} where, in addition, each constraint has assigned a weight.

\begin{definition}
Given a WAF $\Upsilon = \<\A,\R,\C,\W\>$,
a $\sigma$-labelling $\L$ for $\<\A,\R\>$ is a 
maximum-cardinality $\sigma$-labelling
for $\Upsilon$ if $\L\models \C$ and,  let $\W_\L \subseteq \W$ be the set of weak constraints in $\W$ that are satisfied by $\L$, there is (i) no $\sigma$-labelling $\L'$ for $\<\A,\R\>$ with $\L'\models \C$ and (ii) $\W_{\L'} \subseteq \W$ 
such that $\L' \models \W_{\L'}$ and $|\W_{\L}| < |\W_{\L'}|$.
\end{definition}

The set of maximum-cardinality $\sigma$-labellings of a WAF 
$\Upsilon$ will be denoted by ${\tt mc\-}\sigma(\Upsilon)$, 
with $\sigma\in \{\tt co,st,pr,sst\}$.

As stated next, counterfactual and semifactual explanations 
one-to-one correspond to maximum-cardinality labellings of 
appropriate WAFs.

\begin{proposition}\label{prop:waf-c}
For any AF $\Lambda=\<\A,\R\>$,  semantics $\sigma\in \{\tt co,st,pr,sst\}$, goal $g\in \A$, and $\sigma$-labelling $\Lab\in\sigma(\Lambda)$, 
\noindent
$\bullet\ {\cal CF}^{\sigma}(g,\Lab)={\tt mc\-}\sigma(\<\A,\R,\{\neg \ell(\L,g)\}, \{\ell(\L,a)\mid a\in\A\setminus\{g\}\}\>)$;\\
\noindent
$\bullet\ {\cal SF}^{\sigma}(g,\Lab)={\tt mc\-}\sigma(\<\A,\R,\{\ell(\L,g)\}, \{\neg \ell(\L,a)\mid a\in\A\setminus\{g\}\}\>)$.
\end{proposition}

\begin{example}\label{ex:semi-waf}\styleex 
Consider the AF $\<\A,\R\>$  of Example~\ref{running-example} and recall from Example~\ref{example:counter1} that 
${\cal CF}^{\st}(\meat,\L_3)\=\{\L_2\}$.
We have that $\L_2\in$ ${\tt mc}\-\st(\<\A,\R,\C,\W\>)$ where
$\C$ $=$ $\{\neg\i(\meat)\}$,  and  
$\W=\{\o(\fish),$ $\o(\white),$ $\i(\red)\}$
($\C$ consists of a single strong constraint, while $\W$ consists of three weak constraints).
Moreover, recalling from Example~\ref{example:semi1} that 
${\cal SF}^{\st}(\red,$ $\L_3)$ $=$ $\{\L_2\}$, we have that $\L_2\in$ ${\tt mc}\-\st(\<\A,\R,\C,\W\>)$ where 
$\C=\{\i(\red)\}$,  and 
$\W=\{\neg\o(\fish),$ $\neg\o(\white),$ $\neg\i(\red)\}$.~\hfill$\Box$
\end{example}

\subsection*{\textit{asprin} Encoding}

Given the tight relationship between formal argumentation and Answer Set Programming (ASP),  we introduce \textsc{Explain} 
in Algorithm~\ref{algo:Generic-Extension} that 
computes the set of counterfactual and semifactual explanations 
by leveraging existing ASP-based solvers.
In particular, we rely on the \textit{asprin} framework, that is \textit{ASP for preference handling}~\cite{BrewkaD0S15a,BrewkaD0S15b}.
Intuitively, Algorithm~\ref{algo:Generic-Extension} 
encodes the distance measure $\delta$ between labellings/extensions through (weighted) preferences in ASP to 
select best extensions, among extensions given by ASP encodings of AF semantics.
We use $P_{\S}$ to denote a set of rules corresponding to the encoding of semantics $\S$.
As an example, an encoding for stable semantics  is 
as follows~\cite{DvorakGRWW20}: 
\[
  P_{\st} = \left\{  \begin{array}{l}
\textbf{in}(X) \text{ :- } \textbf{not out}(X),  \textbf{arg}(X);\\
\textbf{out}(X)\text{ :- }\textbf{not in}(X), \textbf{arg}(X);\\
\textbf{defeated}(X)\text{ :- }\textbf{in}(Y), \textbf{att}(Y,X);\\
\text{ :- }\textbf{in}(X), \textbf{in}(Y), \textbf{att}(X,Y);\\
\text{ :- }\textbf{out}(X), \textbf{not defeated}(X);
  \end{array}\right\}.
\]

Algorithm~\ref{algo:Generic-Extension} takes as input an AF $\Lambda=\langle \A, \R\rangle$, a goal argument $g\in\A$, a semantics $\sigma\in\{\tt co,st,pr,sst\}$, a $\S$-labelling $\Lab$, and the task type $T \in \{CF,SF\}$ (either counterfactual or semifactual).
After defining the set of ASP rules, $P_{\S}$, 
encoding AF semantics (Line \ref{alg1:sigma-enc}),
the ASP encoding $P_\Lambda$ for AF $\Lambda$ is computed (Line~\ref{alg1:af-enc}).
Then, using the result of Proposition~\ref{prop:waf-c},  
a set $P_S$ containing a single strong constraint (Line \ref{alg1:SC1-enc} for $CF$,  
Line \ref{alg1:SC2-enc} for $SF$) 
and a set of weak constraints $P_W$ (Line \ref{alg1:WC1-enc} for 
$CF$, Line \ref{alg1:WC2-enc} for $SF$) are computed.
As an example, if $T=CF$, to compute {answer sets} representing counterfactuals, the constraint ``:- $\ell(\L,g)$'', stating that $\ell(\L,g))$ must be false is added. 
In contrast, if $T=SF$, to compute {answer sets} representing semifactuals, the constraint ``:- \textbf{not} $\ell(\L,g))$'', stating that $\ell(\L,g))$ must be true, is added.  
Moreover, to ensure that an {answer set} 
corresponds to a counterfactual of $\L$ w.r.t.  $g$, it should satisfy as less constraints of the form  \textbf{w}(a) :- \textbf{not} $\ell(\L,a)$ as possible, 
so that the distance w.r.t. $\L$ is minimized  
(a similar approach is used for semifactuals).
To this end the following \textit{asprin} preference statement and optimization directive are given when invoking \textit{asprin} over $P = P_\sigma \cup P_\Lambda \cup P_S \cup P_W$ (Line~\ref{line:return3}).
\begin{itemize}
\item 
\#preference(p,less(cardinality))$\{\textbf{w}(X)$ : \textbf{arg}$(X)\}$;
\item
\#optimize(p).
\end{itemize}

As stated next, our algorithm is sound and complete.

\begin{theorem}\label{thm:algo}
Algorithm~\ref{algo:Generic-Extension} is sound and complete.
\end{theorem}

\renewcommand{\algorithmicrequire}{\textbf{Input:}} 
\renewcommand{\algorithmicensure}{\textbf{Output:}}

\floatname{algorithm}{Algorithm}
\algsetup{indent=2em}
\begin{algorithm}[t]
\setstretch{.8}
\caption{\textsc{Explain}($\Lambda, g,\sigma, \Lab,T)$}
\begin{algorithmic}[1]
\REQUIRE 
AF $\Lambda=\langle \A, \R\rangle$,  goal $g\in\A$,  
$\sigma\in\{\tt co,st,pr,sst\}$, 
labelling $\Lab\in\sigma(\Lambda)$,
task $T\in \{CF,SF\}$.
\ENSURE 
${\cal CF}^{\sigma}(g,\Lab)$ if $T=CF$, 
${\cal SF}^{\sigma}(g,\Lab)$ if $T=SF$. 
\STATE{Let $P_\sigma$ be the ASP encoding for semantics $\sigma$;}\label{alg1:sigma-enc}
\STATE{$P_{\Lambda}=\{\textbf{arg}(a)\mid a\in \A\}\cup\{\textbf{att}(a,b)\mid (a,b)\in \R\}$;}\label{alg1:af-enc}
\IF{$T= CF$}
	\STATE{\hspace*{-4mm}$P_{S}= \{$:- $\ell(\L,g))\}$;}\label{alg1:SC1-enc}
	\STATE{\hspace*{-4mm}$P_{W}= \{\textbf{w}(a)$:- \textbf{not} $\ell(\L,a)\!\mid\! a\in \A\!\setminus\!\{g\}\}$;}\label{alg1:WC1-enc}
\ELSE
	\STATE{\hspace*{-4mm}$P_{S}= \{$:- \textbf{not} $\ell(\L,g)\}$;}\label{alg1:SC2-enc}
	\STATE{\hspace*{-4mm}$P_{W}= \{\textbf{w}(a)$:- $\ell(\L,a)\!\mid\! a\in \A\!\setminus\!\{g\}\}$;}\label{alg1:WC2-enc}
\ENDIF
\RETURN{\textsc{asprin}($P_\Lambda\cup P_\sigma\cup P_S \cup P_W$);} \label{line:return3}
\end{algorithmic}
\label{algo:Generic-Extension}
\end{algorithm}

\begin{example}\label{ex:algorithm-asp}\rm
Considering the AF $\Lambda$ of Example~\ref{running-example},  $\ST$-labelling $\L_3 = \{\o(\fish), $ $\i(\meat), $ $\o(\white),$ $\i(\red)\}$, and goal argument $\meat$,   the \textit{asprin} program $P$ built in Algorithm~\ref{algo:Generic-Extension} with $T=CF$  is 
$P = P_{\ST} \cup P_{\Lambda} \cup P_{S} \cup P_{W}$ 
where $P_{\ST}$ is as shown earlier and :
\[
  P_{\Lambda} = \left\{\!\! \begin{array}{l}
\textbf{arg}(\fish); \ \ \textbf{arg}(\meat); \ \ \ \ \ \ \ \ \  \textbf{att}(\fish,\meat);\\
\textbf{arg}(\white); \textbf{att}(\meat,\fish); \textbf{att}(\meat,\white);\\
\textbf{arg}(\red);\ \ \ \ \textbf{att}(\red,\white); \textbf{att}(\white,\red);
  \end{array}\!\!\right\}
\]
\[
  P_{S} =\ \{\ \text{:- } \textbf{in}(\meat);\ \}\hspace*{54mm} \\ 
\]
\[
  P_{W} = \left\{\!\! 
\begin{array}{l}
\textbf{w}(\fish)\ \  \text{ :- }  \textbf{not out}(\fish);\\
\textbf{w}(\white) \text{ :- } \textbf{not out}(\white);\\
\textbf{w}(\red)\ \ \ \, \, \text{ :- } \textbf{not in}(\red);\\
\end{array}\!\!\right\}\hspace*{30mm}
\]

The result of \textsc{asprin}($P$) is ${\cal CF}^{\st}(\meat,\L_3)=\{\L_2 \}$.  ~\hfill{$\Box$}
\end{example}

\section{{Discussion}}\label{sec:alternative}

In this section, we consider definitions of counterfactual and semifactual more general than those in given 
Definitions~\ref{def:counter1} and~\ref{def:semi1}, respectively.
In particular, we focus on the following extensions: 
(\textit{a}) more general distance measures between labellings;
(\textit{b}) more general criterion for changing the status of the goal argument; and 
\textit{(c)} set of goal arguments (instead of a single argument).
Notably, regardless of the above-mentioned generalization  
adopted for Definitions~\ref{def:counter1} and~\ref{def:semi1}, 
the complexity bounds in Table~\ref{tab:my-table2} still hold.

\paragraph{More general distance measures.}
The measure $\delta$  used Definitions~\ref{def:counter1} and~\ref{def:semi1} does 
not distinguish among different labelling changes.  
For example, labelling $\i({\fish})$ has the same distance, $1$, from $\o(\fish)$ or $\un(\fish)$.
Here,  we redefine  the concept of counterfactual and semifactual w.r.t. a more general distance measure $\eta$, that is, a (polynomial-time computable) function $\eta: \A\times\{\i,\o,\un\}\times\{\i,\o,\un\}\rightarrow \mathbb{N}$,
which assigns a (positive) number to every possible labelling change for each argument. 
Then, we use $\eta(\Lab,\Lab')=\sum_{a\in \A} \eta(a,\L(a),\L'(a))$ to denote the (weighted) 
distance between $\sigma$-labellings $\L$ and $\L'$. 

Observe that $\delta$ is a special case of $\eta$ where, 
$\forall a\in\A$,  $\eta(a,\L(a), \L'(a))= 0$ if $\L(a)=\L'(a)$, $1$ otherwise.

Given the new distance measure $\eta$, we can redefine 
counterfactuals and semifactuals by replacing item \textit{(ii)} of
Definitions~\ref{def:counter1} and~\ref{def:semi1} as follows.
\begin{itemize}[leftmargin=20pt]
\item
Minimality condition for Definition~\ref{def:counter1}:
there is no $\Lab'' \in \sigma(\Lambda)$ s.t.   $\Lab(g)\neq\Lab''(g)$ and $\eta(\Lab,\Lab'')<\eta(\Lab,\Lab')$.
\item
Maximality condition for Definition~\ref{def:semi1}:
 there is no $\Lab'' \in \sigma(\Lambda)$ s.t.   $\Lab(g)=\Lab''(g)$ and $\eta(\Lab,\Lab'')>\eta(\Lab,\Lab')$.
\end{itemize}

As a proposal for $\eta$ to capture the idea that 
the distance should distinguish between the changes $\i(a)$-to/from-$\o(a)$ from those $\i(a)$-to/from-$\un(a)$ or $\o(a)$-to/from-$\un(a)$,  
we could define it as follows:
$\eta(a,\L(a), \L'(a))=0$ if $\L(a)=\L'(a)$;
otherwise,  
$\eta(a,\L(a), \L'(a))=1$ (resp.,  $2$) if $\un(a)\in\{\L(a),\L'(a)\}$ (resp.,  $\un(a)\not\in\{\L(a),\L'(a)\}$).
This way changes between decided values are weighted double of the others.

\begin{example}\styleex
Consider the AF $\Lambda$ shown in Figure~\ref{fig:intro} whose $\co$-labellings are :\\
\noindent
$\L_1 = \{\i(\fish), \o(\meat), \i(\white),\o(\red)\}$, \\
\noindent
$\L_2 = \{\i(\fish), \o(\meat), \o(\white),\i(\red)\}$, \\
\noindent 
$\L_3 = \{\o(\fish), \i(\meat), \o(\white),\i(\red)\}$, \\
\noindent
$\L_4 = \{\un(\fish), \un(\meat), \o(\white),\i(\red)\}$, \\
\noindent
$\L_5 = \{\i(\fish), \o(\meat), \un(\white),\un(\red)\}$, \\
\noindent 
$\L_6 = \{\un(\fish), \un(\meat), \un(\white),\un(\red)\}$.

Using distance $\delta$, we have that
${\cal CF}^{\co}(\meat,\L_3)=\{\L_2,$ $\L_5\}$ and ${\cal SF}^{\co}(\red,\L_3)$ $=$ $\{\L_2,\L_4\}$, 
while for $\eta$ defined as in the paragraph preceding this example,
we have that ${\cal CF}^{\co}(\meat,\L_3)$ $=$ $\{\L_5\}$ and ${\cal SF}^{\co}(\red,\L_3)$ $=$ $\{\L_2\}$.~\hfill~$\Box$
\end{example}

As stated next, our complexity results holds irrespective 
of how $\eta$ is instantiated.

\begin{proposition}\label{prop:alternative-distance}
The results of Theorems~\ref{thm:kexc}--\ref{theorem:semi-sa} still hold if measure $\eta$ is used (instead of $\delta$)
in Definitions~\ref{def:counter1} and~\ref{def:semi1}.

\end{proposition}

\paragraph{Replacing criterion for goals' status change.}
A counterfactual $\L'$ for $\L$ and $g$ could also be defined by requiring that the status of the goal $g$ w.r.t. $\L'$ is not undecided, that is by changing condition $(i)$ of Definition~\ref{def:counter1}. 
For instance, we could replace condition $(i)\ \Lab(g)\neq\Lab'(g)$,  
equivalently rewritten as
\begin{tabular}{cc}
\begin{tabular}{c}
$\L'(g)\in\Bigg\{$
\end{tabular}
&
\hspace*{-8mm}
\begin{tabular}{ll}
$\{\o,\un\}$&$\text{if}\ \L(g)=\i$\\
$\{\i,\un\}$&$\text{if}\ \L(g)=\o$\\
$\{\i,\o\}$&$\text{if}\ \L(g)=\un$
\end{tabular}
\end{tabular}
 
\noindent 
with the following one:

\hspace*{-4.5mm}
\begin{tabular}{cc}
\begin{tabular}{c}
$\L'(g)\in\Bigg\{$
\end{tabular}
&
\hspace*{-8mm}
\begin{tabular}{ll}
$\{\o\}$& $\ \ \ \ \text{if}\ \L(g)=\i$\\
$\{\i\}$&$\ \ \ \ \text{if}\ \L(g)=\o$\\
$\{\i,\o\}$&$\ \ \ \ \text{if}\ \L(g)=\un$
\end{tabular}
\end{tabular}

Again, replacing condition $(i)$ of Definition~\ref{def:counter1}
with that above---or any other that can be checked in PTIME---does not alter the complexity results in Table~\ref{tab:my-table2}.

\paragraph{Considering multiple goal arguments.}
For the sake of presentation,  we focused on a single goal argument $g\in\A$.
However, a set $S\subseteq\A$ of goal arguments whose status 
is required to change (resp.,  to not change) could be considered in Definitions~\ref{def:counter1} and~\ref{def:semi1}, respectively,
or their generalizations introduced earlier.
More formally, given an AF $\Lambda=\<\A,\R\>$,  distance measure $\eta$,  semantics $\sigma\in\{\tt co,st,pr,sst\}$,  goal arguments $S\subseteq \A$, and $\sigma$-labelling $\Lab$,  
we say that $\Lab'$ is an $\eta$-counterfactual 
(resp.,  $\eta$-semifactual) of $\Lab$ w.r.t.  $S$ 
if condition $(i)$ and $(ii)$ of Definition~\ref{def:counter1} (resp., Definition~\ref{def:semi1}) are satisfied for all $g\in S$,
where measure $\eta$ is used (instead of $\delta$).  
Still, adopting these definitions of counterfactuals and semifactuals does not alter the complexity bounds of Theorems \ref{thm:kexc}--\ref{theorem:semi-sa} 
as conditions $(i-ii)$ can be checked in PTIME.

\section{Related Work}\label{sec:rw}
Several researchers explored how to deal with explanations in formal argumentation.  
Important work 
includes e.g.~\cite{FanT15}, where a new argumentation semantics is proposed for capturing explanations in AF, and~\cite{CravenT16} that focuses on ABA framework{s}~\cite{CravenT16,DungKT09,Hung16}. 
They treat an explanation as a semantics to answer why an argument is accepted or not. 
In~\cite{FanT15} an explanation is as a set of arguments justifying a given argument by means of a proponent-opponent dispute-tree~\cite{DungMT07}. 
An 
approach based on debate trees as proof procedure 
for computing grounded, ideal, and preferred semantics, 
has been proposed in~\cite{ThangDH09}. 
The approach in~\cite{AlfanoCGPT23} build explanations that are sequences of arguments by exploiting topological dependencies among arguments.
An alternative definition for explaining complete extensions has been proposed in~\cite{Baumann021}. 
It exploits the concept of \textit{reduct}, i.e. a sub-framework obtained by removing true and false arguments w.r.t.  a complete extension.
The concept of \textit{strong explanation} is proposed in~\cite{0001W21a}, inspired by the related notions introduced in~\cite{BrewkaU19,BrewkaTU19,SaribaturWW20}.

Counterfactual reasoning in AF has been firstly introduced in~\cite{sakama2014counterfactual}, where considering 
sentences of the form ``if $a$ were rejected, then $b$ would be accepted'',  an AF $\Lambda$ is modified to another AF $\Lambda'$ such that (\textit{i}) argument $a$ which is accepted in $\Lambda$ is rejected in $\Lambda'$ (\textit{ii}) and the $\Lambda'$ is as close as possible to $\Lambda'$. 
An interesting problem related to this
is \textit{enforcement}~\cite{NiskanenWJ16,NiskanenWJ18,WallnerNJ17},  
that is how changes in the AF affects the acceptability of arguments, and how to modify an AF 
to guarantee that some arguments get a given labelling.   
In particular, 
extension enforcement concerns how to modify an AF to ensure that a given set of arguments becomes (part of) an extension~\cite{BaumannDMW21}. 
Moreover, in~\cite{borg2024minimality}, a framework for determining argument-based explanations in both abstract and structured settings is proposed. 
The framework is able to answer why-question, such as `why is argument $a$ credulously accepted under $\pr$'?
Finally, an approach to explain the relative change in strength of specified arguments in a Quantitative Bipolar AF updated by changing its arguments, their initial strengths and/or relationships has been recently proposed in \cite{KampikCA24}.

However, none of the above-mentioned approaches deals with semifactual reasoning and most of them manipulate the AF by adding  arguments or meta-knowledge.
In contrast, in our approach, focusing on a given AF, novel definitions of counterfactual and semifactual are introduced to help understand what should be different in a solution (not in the AF) to accommodate a user requirement concerning a given goal.

\section{Conclusions and Future Work}
We have proposed the concept of counterfactual and semifactual explanations in abstract argumentation, and investigated the complexity of counterfactual- and semifactual-based reasoning in AF. 
It turns out that the complexity of the considered problems 
is not lower than those of corresponding classical problems in AF, and is provably higher for fundamental problems 
such as the verification problem. 
It is worth mentioning that, though their formulation is similar and they share the same complexity bounds, the considered counterfactual-based and semifactual-based problems are not dual problems---we do not see how to naturally reduce one to the other; however, a (possibly complex) reduction may exist as our complexity results do not rule this out. 

We have provided ways for computing semifactuals and counterfactuals by reducing to weak-constrained AF and 
then to ASP with preferences, enabling implementations
by using \textit{asprin}~\cite{BrewkaD0S15b}.
It can be shown that our complexity results carry over to other frameworks whose complexity is as that of AF,
such as Bipolar AF~\cite{CohenGGS14} and AF with recursive attacks and supports~\cite{CohenGGS15,CayrolFCL18}, 
among others~\cite{VillataBGT12,Gottifredi-Cohen-Garcia-Simari18,DvorakKUW24}. 

Future work will be devoted to considering 
more general forms of AF, 
such as incomplete and probabilistic AF~\cite{LiON11,Hunter12,Hunter13,AIJ2021} as well as structured argumentation formalisms~
\cite{ModgilP14,CyrasHT21,GarciaPS20}.

\balance 
\bibliographystyle{stylerefs}
\bibliography{refs}

\end{document}